\documentclass{article}

\usepackage{microtype}
\usepackage{graphicx}
\usepackage{subcaption}

\usepackage{float}
\usepackage{booktabs} % for professional tables

\usepackage{amsmath}
\usepackage{multirow}
\usepackage{hyperref}
\usepackage{makecell}

\usepackage[accepted]{icml2021}

\icmltitlerunning{Siamese Labels Auxiliary Learning}
\begin{document}

\twocolumn[
\icmltitle{Siamese Labels Auxiliary Learning}
%\icmltitle{Siamese Labels Auxiliary Network (SiLaNet)}

% It is OKAY to include author information, even for blind
% submissions: the style file will automatically remove it for you
% unless you've provided the [accepted] option to the icml2021
% package.

% List of affiliations: The first argument should be a (short)
% identifier you will use later to specify author affiliations
% Academic affiliations should list Department, University, City, Region, Country
% Industry affiliations should list Company, City, Region, Country

% You can specify symbols, otherwise they are numbered in order.
% Ideally, you should not use this facility. Affiliations will be numbered
% in order of appearance and this is the preferred way.
\icmlsetsymbol{equal}{*}

\begin{icmlauthorlist}
\icmlauthor{Wenrui Gan}{}
\icmlauthor{Zhulin Liu}{}
\icmlauthor{C. L. Philip Chen}{}
\icmlauthor{Tong Zhang}{}
%\icmlauthor{Fiuea Rrrr}{to}
%\icmlauthor{Tateu H.~Yasehe}{ed,to,goo}
%\icmlauthor{Aaoeu Iasoh}{goo}
%\icmlauthor{Buiui Eueu}{ed}
%\icmlauthor{Aeuia Zzzz}{ed}
%\icmlauthor{Bieea C.~Yyyy}{to,goo}
%\icmlauthor{Teoau Xxxx}{ed}
%\icmlauthor{Eee Pppp}{ed}
\end{icmlauthorlist}
%
%\icmlaffiliation{}{}
%\icmlaffiliation{goo}{Googol ShallowMind, New London, Michigan, USA}
%\icmlaffiliation{ed}{School of Computation, University of Edenborrow, Edenborrow, United Kingdom}
%
%\icmlcorrespondingauthor{Tong Zhang}{tony@scut.edu.cn}
%\icmlcorrespondingauthor{Eee Pppp}{ep@eden.co.uk}

% You may provide any keywords that you
% find helpful for describing your paper; these are used to populate
% the "keywords" metadata in the PDF but will not be shown in the document
\icmlkeywords{Siamese Labels, Auxiliary training, Parameter compression, Classification task}

\vskip 0.3in
]

% this must go after the closing bracket ] following \twocolumn[ ...

% This command actually creates the footnote in the first column
% listing the affiliations and the copyright notice.
% The command takes one argument, which is text to display at the start of the footnote.
% The \icmlEqualContribution command is standard text for equal contribution.
% Remove it (just {}) if you do not need this facility.

\printAffiliationsAndNotice{}  % leave blank if no need to mention equal contribution
%\printAffiliationsAndNotice{\icmlEqualContribution} % otherwise use the standard text.

\begin{abstract}
In deep learning, auxiliary training has been widely used to assist the training of models. During the training phase, using auxiliary modules to assist training can improve the performance of the model. During the testing phase, auxiliary modules can be removed, so the test parameters are not increased. In this paper, we propose a novel auxiliary training method, Siamese Labels Auxiliary Learning (SiLa). Unlike Deep Mutual Learning (DML), SiLa emphasizes auxiliary learning and can be easily combined with DML. In general, the main work of this paper include: (1) propose SiLa Learning, which improves the performance of common models without increasing test parameters; (2) compares SiLa with DML and proves that SiLa can improve the generalization of the model; (3) SiLa is applied to Dynamic Neural Networks, and proved that SiLa can be used for various types of network structures.
\end{abstract}

\section{Introduction}
\label{introduction}

Since the popularity of convolutional neural networks  \cite{krizhevsky2012imagenet,lecun1989backpropagation}, people have designed many types of convolutional neural network models. Such as VGGNet \cite{simonyan2014very}, GoogleNet \cite{Szegedy_2015_CVPR}, ResNet \cite{he2016deep}, etc. All of these models have made important contributions to the development of deep learning. VGGNet \cite{simonyan2014very} was designed to have deeper convolution layer to improve model representation capabilities. On the other hand, GoogleNet \cite{Szegedy_2015_CVPR} was designed from another perspective. It consists of well-designed inception module, and a large number of $1*1$ convolution kernels, which were used in GoogleNet for dimensional increasing or reducing operations. Additionally, multiple scale convolution kernels are used simultaneously in inception module. Such kinds of improvements have enhanced the representation capabilities of GoogLeNet. Then, Kaiming He et al. designed the Residual Network \yrcite{he2016deep} with shortcut connections structure \cite{bishop1995neural,ripley2007pattern,venables2013modern}. The residual structure alleviates the vanishing/exploding gradients problem \cite{bengio1994learning,glorot2010understanding} of deep neural networks, and makes it possible for people to design deeper network structures.

As the model gets deeper and deeper, the parameter size of the model becomes larger and larger, which makes it more and more difficult to run the model on some resource-constrained devices, such as mobile devices. It has become a hot research to improve the performance of the model as much as possible without increasing or even reducing the amount of computation. There are many ways to achieve this, including Auxiliary Training \cite{lee2015deeply,Szegedy_2015_CVPR}, Knowledge Distillation \cite{bucilua2006model,hinton2015distilling}, Dynamic Neural Networks \cite{huang2018multi,han2021dynamic}, etc. Among them, Auxiliary Training uses additional auxiliary modules in the training phase and removes auxiliary modules in the testing phase. Knowledge Distillation can be understood as using a well-trained teacher model with a larger number of parameters to assist the training of the student model. On the basis of Knowledge Distillation, Deep Mutual Learning \cite{lee2015deeply} is proposed. It removes the teacher model and uses only a set of student models to learn from each other.

On the basis of these studies, this paper proposes an auxiliary training method. It also uses a set of student models for mutual auxiliary training, which ultimately improves the performance of the model without increasing the amount of test computation. It should be noted that SiLa is a new learning method completely different from DML, and the two methods can be easily combined. First, based on the analysis of the cross-entropy loss function, we propose the siamese label, and design the basic structure of the Siamese Labels Auxiliary (SiLa) Module on the basis of the Siamese Label. SiLa Module IS designed to be used in the network structures trained using auxiliary modules. By using SiLa, the various models are more closely linked, and the auxiliary information can be better passed between the models, which ultimately improves the performance.

Specifically, we concatenate the outputs of the two models to form Siamese Labels. Since the Siamese Labels are actually the outputs of different models corresponding to the same sample, there is a one-to-one correspondence within the Siamese Labels. Then, the Siamese Labels are input to the cross-entropy loss function of each model, and the loss of each model will carry the output information from other models, and use this information as auxiliary information to assist its own training. At the same time, these auxiliary information will be transmitted back to other models through back-propagation, and other models will be adjusted accordingly. For detailed analysis, see Section~\ref{Siamese_labels}.

\section{Related Work}
\label{related_work}
{\bf Auxiliary Training}\quad Deeply-Supervised Nets (DSN) \cite{lee2015deeply} is a classic paper in the study of training using auxiliary modules. In the training process of DSN, by adding multiple auxiliary classifiers to the hidden layer, the model provides supervised learning not only in the output layer but also in the hidden layer. Through this method, DSN can effectively alleviate the problem of gradient disappearance, and can also make the hidden layer more directly supervised training, and the output layer of the model can be better supervised learning based on the hidden layer. After DSN, the auxiliary classifier modules is used in the network structure of GoogLeNet \cite{Szegedy_2015_CVPR}. Experiments show that using the auxiliary module for auxiliary training can effectively improve the performance of the model. The SiLa proposed in this paper is a more generally applicable method based on these auxiliary training methods. In SiLa, the auxiliary module can be a classifier or an independent network.\\
{\bf Mutual Learning}\quad Deep Mutual Learning (DML) \cite{lee2015deeply} is developed from knowledge distillation \cite{bucilua2006model,hinton2015distilling}. In some respects, DML shares similarities with our proposed SiLa. They both train a set of models at the same time, so that the models influence each other and jointly improve the performance. The difference is that DML uses KL divergence to enable the trained models to learn each other's output distributions, and enables the models to imitate each other to jointly improve performance. The SiLa proposed by us emphasizes auxiliary learning more, and the trained models will assist each other through the SiLa module and make progress together. In addition, SiLa and DML are not mutually exclusive, and the two methods can easily be used simultaneously.\\
{\bf Multi-Scale Dense Convolutional Networks (MSDNet)}\quad  MSDNet \cite{huang2018multi} is an early-exiting dynamic neural network \cite{han2021dynamic}. It uses Multi-scale feature maps and Dense connectivity \cite{huang2017densely}. MSDNet trains multiple classifiers at different depths, and depending on the computing resources, the model uses different classifiers for output during testing. Therefore, under the condition of limited computing resources, the prediction accuracy of the model can be improved as much as possible. Since multiple classifiers of MSDNet are trained simultaneously, SiLa can be used for auxiliary training of these classifiers, thereby improving the performance of each classifier.

\section{Proposed Method}
\label{method}

\subsection{Siamese Labels}
\label{Siamese_labels}

\begin{figure}[bp]
\vskip 0in
%\vskip 0in
\begin{center}
\centerline{\includegraphics[width=\columnwidth]{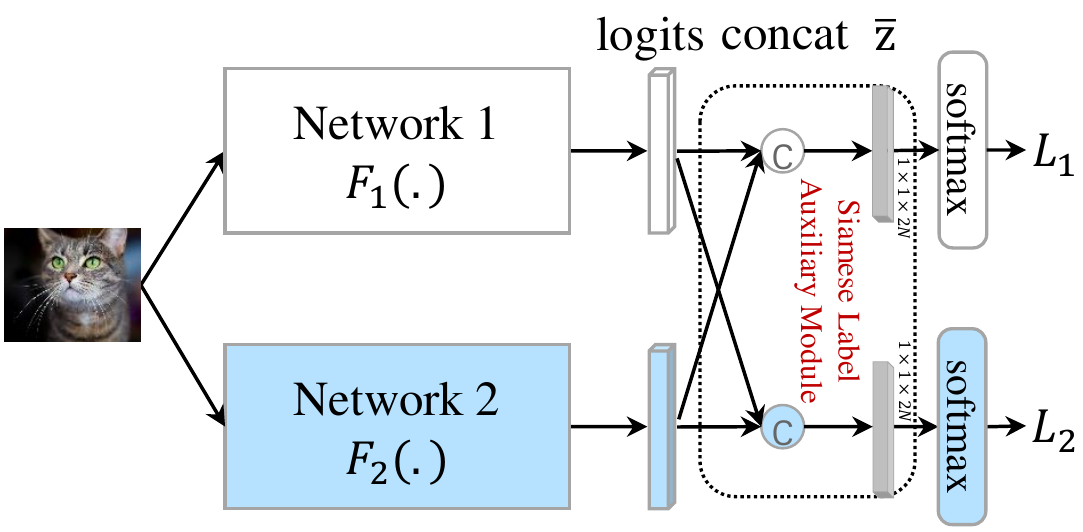}}
\caption{Network structure of a set of independent networks using SiLa for Auxiliary Learning.}
\label{siamese_sila_structure}
\end{center}
\vskip -0.2in
%\vskip 0in
\end{figure}

In traditional $N\raisebox{0mm}{-}classification$ tasks using neural networks, it is customary to map the outputs of the model to N\raisebox{0mm}{-}dimensional vectors at the output layer. These N\raisebox{0mm}{-}dimensional vectors and labels are then input into the cross-entropy loss function to obtain the training loss, and finally the gradient backpropagation is used for optimization. In this paper, we expand the N\raisebox{0mm}{-}dimensional vectors by a factor of $C$ (generally C=2) to obtain C*N\raisebox{0mm}{-}dimensional vectors. This process is achieved by using auxiliary modules. Since the expanded  C*N\raisebox{0mm}{-}dimensional vectors have a one-to-one correspondence, we call them Siamese Labels. The structure that uses the auxiliary module to obtain the Siamese Label is called the Siamese Label Auxiliary Module (SiLa), and the specific structure is shown in Figure~\ref{siamese_sila_structure}. By using the SiLa Module, the model can find more robust parameters as much as possible during the optimization process, thereby improving the performance of the model.

For an N-classification task, given a training set $D= \cup _{i=1}^{K} \{(x_{i},y_{i})\}$ and a base model $F(.)$, where $K$ represents the number of samples, $x_{i}$ represents the i-th sample, and $y_{i}\in \{1,2...N\}$ is a scalar, indicating that the sample $x_{i}$ belongs to the $y_{i}$-th class. $z_{i}=F(x_{i})$ is the output of the base model $F(.)$ using $x_{i}$ as input. And $z_{i}=[z_{i}^{1}, z_{i}^{2}...z_{i}^{y_{i}}...z_{i}^{N}]$, $z_{i}\in R^ {N}$, where $z_{i}^{y_{i}}$ is the $y_{i}$\raisebox{0mm}{-}th component of $z_{i}$. For the convenience of description here, we take a single sample $x$, the corresponding label $y$, and the model output $z$ to illustrate. For the $N\raisebox{0mm}{-}classification$ task, the model is trained using the Cross-Entropy loss function, defined as follows:
\begin{equation}
\begin{aligned}
L_{CE} =\frac{1}{K}\sum_{i=1}^{K}L_{cross}(z_{i},y_{i})
\end{aligned}
\end{equation}
where $L_{cross}$ is the Cross-Entropy loss function for a single sample:
\begin{equation}
\begin{aligned}
L&_{cross}(z,y) =-\sum_{n=1}^{N}I(y,n)log\frac{e^{z^{n} }}{ {\textstyle \sum_{l=1}^{N}e^{z^{l} }}}\\
& = -z^{y}+log(e^{z^{y}} +{ {\textstyle \sum_{l=1,l\ne y}^{N}e^{z^{l} }}} )
\end{aligned}
\end{equation}
where $z^{y}$ is the $y$\raisebox{0mm}{-}th component of $z$,and indicator function $I$ defined as:
\begin{equation}
\begin{aligned}
I( y,n)=\left\{\begin{matrix}1 \quad n=y
 \\0 \quad n\ne y
\end{matrix}\right.
\end{aligned}
\end{equation}
Let $e^{\alpha} =\sum_{l=1,l\ne y}^{N}e^{z^{l} }$, then:
\begin{equation}
\begin{aligned}
L&_{cross}(z,y) =-z^{y}+log(e^{z^{y}} +e^{\alpha}  )
\end{aligned}
\end{equation}
\begin{equation}
\begin{aligned}
\frac{\partial L_{cross}}{\partial z^{y} } &=-\frac{1}{e^{z^{y}-\alpha}+1}\\
\frac{\partial L_{cross}}{\partial \alpha  } &=\frac{1}{e^{z^{y}-\alpha }+1}
\label{lcross-partial}
\end{aligned}
\end{equation}

We know that when training a model using gradient backpropagation, the gradient tends to 0 as the model tends to converge. That is, $\frac{\partial L_{cross}}{\partial z^{y} }$ and $\frac{\partial L_{cross}}{\partial \alpha}$ will tend to 0 as the model tends to converge. It can be seen from Equation ~\ref{lcross-partial} that when $z^{y}$ is larger than $\alpha$, the partial derivatives $\frac{\partial L_{cross}}{\partial z^{y} }$ and $\frac{\partial L_{cross}}{\partial \alpha}$ will quickly become 0, which is not conducive to the back-propagation of the gradient. As is shown in Figure~\ref{siamese_sila_structure}, we concatenate the outputs of the two network models $F_{c}(.)$ ($c\in\left \{ 1,2 \right \}$) as Siamese Labels. When we use Siamese Labels, the number of dimensions increases from $N$ to $2\!*\!N$, since $\alpha =log\textstyle \sum_{l=1,l\ne y}^{N}e^{z^{l} }$, $\alpha$ will increase accordingly. At this time, the Siamese Labels $\bar{z}$ is shown in Equation~\ref{z_bar}, and the losses $L_{1}$ and $L_{2}$ are shown in Equation ~\ref{l_c}. Since the partial derivative $\frac{\partial L_{cross}}{\partial z^{y} }$ tends to 0 when the model converges, when using loss $L_{1}$ for gradient backward, the increase of $\alpha$ makes the value of $z^{y}$ tend to be larger than when the SiLa is not used. But the value of $z^{y}$ does not necessarily tend to be larger than without the use of Siamese Labels, because the value of $z^{y}$ will be reduced when using loss $L_{2}$ for gradient backward. This is exactly what we need, the target output $z^{y}$ of the model is pulled by the forces in two different directions, making the model less likely to fall into local minima. It should be noted that when we use Siamese labels, the added $N$ categories correspond to the $N$ categories in the sample one-to-one. The expanded $2\!N$ dimensions are divided into $2$ groups with each $N$ dimensions as a group. After expansion, $\bar{z}=[z^{1}...z^{y}...z^{N}...z^{N+1}...z^{N+y}...z^{2N}]$, $\bar{z}\in R^ {2N}$. As shown in Figure~\ref{siamese_sila_structure}, the corresponding classification loss of each group is $L{1}$ and $L_{2}$, and its corresponding weight value is $\beta_{c}$ ($c\in\left \{ 1,2 \right \}$). $\beta_{c} $ is the hyperparameter that needs to be set. The corresponding SiLa loss $L_{sila}$ is shown in Equation~\ref{sila-loss1}. For the training process, see Algorithm~1.
\begin{table}[h]
\begin{tabular}{p{.99\linewidth}}
\toprule
\textbf{Algorithm 1:  Siamese Labels Auxiliary Learning  }\\
\midrule
\textbf{\emph{Input: }} Training set $\mathcal{X}$, label set $\mathcal{Y}$, learning rate $\gamma_{t}$, loss weight $\beta_{1}$ and $\beta_{2}$ \\ 
\textbf{\emph{Initialize: }} Models $\Theta_{1}$ and $\Theta_{2}$ to different initial conditions.  \\
\textbf{\emph{Repeat :}} \\
\quad  $t = t + 1$ \\
\quad  Randomly sample data $\boldsymbol{x}$ from $\mathcal{X}$. \\
\quad  \textbf{1: }  Compute Siamese Labels $\bar{z}$ by~\ref{z_bar}.\\
\quad  \textbf{2: }  Compute Loss $L_{sila}$ by~\ref{sila-loss1}.\\
\quad  \textbf{3: } Compute the stochastic gradient and update $\Theta_{1}$, $\Theta_{2}$: \\
\begin{equation}
\Theta_{1} \leftarrow \Theta_{1} + \gamma_{t} \frac{\partial L_{sila}}{\partial \Theta_{1} }
\end{equation}
\begin{equation}
 \Theta_{2} \leftarrow \Theta_{2} + \gamma_{t} \frac{\partial L_{sila}}{\partial \Theta_{2} }
\end{equation}
\textbf{\emph{Until :}}  convergence \\
\bottomrule
\end{tabular}
\vspace{-2mm}
\nonumber
\end{table}

%The weight of the first group in the cross entropy is $1$, and the weight of the other group in the cross entropy is $\beta_{c}$ $2\le c \le C$. $\beta $ is the hyperparameter that needs to be set. The pytorch code corresponding to the SiLa loss function is shown in Figure~\ref{pytorch_code}. The corresponding SiLa loss function $L_{sila}$ is shown in Equation~\ref{sila-loss1}.
\begin{equation}
\begin{aligned}
L_{c}  = \frac{1}{K}\sum_{i=1}^{K}L_{cross}(\bar{z}_{i},y_{i}+(c-1)N)~~~~c\in\left \{ 1,2 \right \}
\label{l_c}
\end{aligned}
\end{equation}
where $\bar{z}$ is obtained by concatenating the outputs $F_{c}$ of the two models:
\begin{equation}
\begin{aligned}
\bar{z}_{i} =concat(\left [ F_{1}(x_{i}),F_{2}(x_{i}) \right ] ,dim = 1)
\label{z_bar}
\end{aligned}
\end{equation}
The total loss $L_{sila}$ is:
\begin{equation}
\begin{aligned}
L_{sila} &=\sum_{c=1}^{2}\beta_{c}L_{c}\\
&=\frac{1}{K}\sum_{i=1}^{K}\sum_{c=1}^{2}\beta_{c}L_{cross}({\bar{z}_{i}},y_{i}+(c-1)N)
\label{sila-loss1}
\end{aligned}
\end{equation}
In this subsection, we introduce the Siamese Label Auxiliary Module and prove its logical feasibility with the formula. By using the Siamese Label auxiliary module, the model is less likely to get stuck in local minima during training. This improves the classification performance of the model.

%\subsection{Use the Siamese Label Auxiliary module in the training of common models}
%\subsection{Use the SiLa module in the training of common models}
%\subsection{Using Siamese Label Auxiliary Module in Collaborative Learning}
\subsection{Siamese Labels Auxiliary Learning (SiLa)}
\label{collaborative_learning}

In this section, in order to better study the Siamese Label Auxiliary Module, we use the network structure shown in Figure~\ref{siamese_sila_structure}, where Network 1 and Network 2 are two independent network models, which can be ResNet \cite{he2016deep}, VGG \cite{simonyan2014very}, DenseNet \cite{huang2017densely} et al. We concatenate the logits output by Network 1 and Network 2 with the Siamese Label Auxiliary Module to get the Siamese label $\bar{z}$. Then, the Siamese Labels $\bar{z}$ and sample labels $y$ are input to the SiLa loss function to obtain the loss $L_{sila}$ required for model training. Similar to deep mutual learning (DML) \cite{zhang2018deep}, our proposed Siamese Label Auxiliary Module enables Network 1 and Network 2 to assist each other in the learning process, and ultimately both can achieve better performance than independent learning. Different from DML, DML uses KL divergence to make models learn from each other's distribution, focusing more on the concept of mutual learning. The Siamese Label Auxiliary Module we proposed is to assist each other in the model learning process, focusing more on the concept of auxiliary learning.

As shown in Figure~\ref{siamese_sila_structure}, in our network model, during the training phase, both  Network 1 and Network 2 participate in the training. These two networks can assist each other and become each other's auxiliary models. But in the testing phase, we can choose only one of the models for testing. In this paper, we call the network structure composed of multiple output nodes and Siamese Label Auxiliary Module as Siamese Labels Auxiliary Network (SiLaNet). Different from the teacher model and the student model in Knowledge Distillation, only the student model exists in our network, they assist each other and make progress together.

It should be noted that the SiLa does not limit the composition of the network structure, we can try to design different auxiliary modules, as long as their outputs meet the input conditions of SiLa loss. In the next section we will explore the use of SiLa in a network structure with multiple output nodes.

\begin{figure}[htbp]
%\vskip 0in
\vskip 0in
\begin{center}
\centerline{\includegraphics[width=\columnwidth]{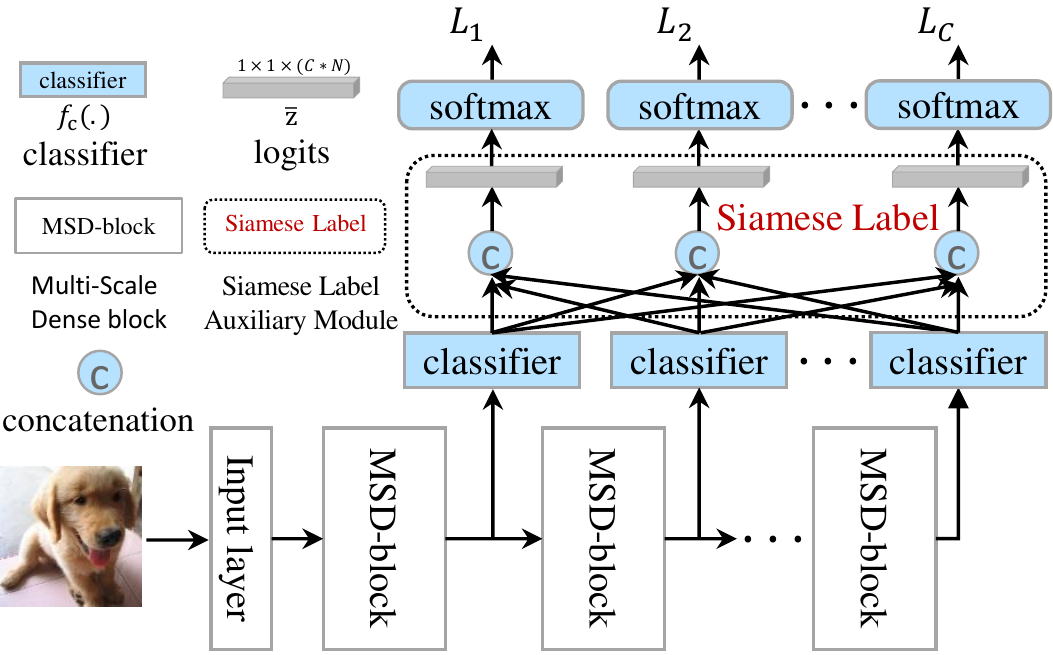}}
\caption{Early-exiting Dynamic Neural Network architecture using SiLa for Auxiliary Learning based on MSDNet.}
\label{SilaNet_figure}
\end{center}
\vskip -0.2in
\end{figure}

\subsection{Dynamic SiLa}
%\subsection{Using SiLa Module in Dynamic Neural Network}
\label{msd_sila}
According to the analysis in Section~\ref{Siamese_labels}, the use of the Siamese Label auxiliary Module requires the network structure to have multiple output nodes. In a Dynamic Neural Network (DNN) using early-exiting schemes, there happens to be multiple classification output nodes. In this subsection, we use the Siamese Labels Auxiliary Module in a dynamic neural network using an early-exiting schemes, and the required network structure on the basis of Multi-Scale DenseNet (MSDNet) \cite{huang2018multi}. The specific structure is shown in Figure~\ref{SilaNet_figure}. The MSD-block in the structure is the Multi-Scale Dense structure designed by Huang G et al.\cite{huang2018multi}. It uses Multi-scale feature maps and Dense connectivity. After each  MSD-block structure, a classifier $f_{c}(.)$ is used for output, from shallow to deep, there are a total of $C$ blocks and classifiers. We directly concatenate the output of this $C$ classifier through the Siamese Label Auxiliary Module to get Siamese Labels $\bar{z}$, that is, $\bar{z}=concat(f_{1}...f_{C})$, which constitutes the Siamese Label Auxiliary Module. Then input the Siamese Labels $\bar{z}$ and sample labels $y$ into the SiLa loss function to obtain the loss $L_{msd\_sila}$ required for model training, see Equation~\ref{msdsila} for details. 
\begin{equation}
\begin{aligned}
L_{msd\_sila} &=\sum_{c=1}^{C}\beta_{c}*L_{c}
\label{msdsila}
\end{aligned}
\end{equation}
As shown in Figure~\ref{SilaNet_figure}, in our network model, in the training phase, the outputs of $C$ classifiers all participate in the training of the network. By using the SiLa Module, each classifier assists other classifiers, and each classifier is assisted by other classifiers. That is, the outputs of the $C$ classifiers assist each other and make progress together, which improves the performance of the model.

In the testing phase of the model, like the MSDNet, the model uses different classifiers for output according to the computing resources and the difficulty of classifying samples. In the Anytime Prediction task \cite{grubb2012speedboost}, a fixed amount of computing resources is used for each sample, and the model outputs in the classifier that is closest to fully using computing resources. In the Budgeted Batch Classification task \cite{huang2018multi}, a total amount of computing resources is fixed for a batch of samples to be processed, in which the easy-to-classify samples are output from the shallow classifier, and the difficult-to-classify samples are output from the deep classifier. Since the Siamese Label Auxiliary Module is used in the training process of the model, the output performance of each classifier has been improved. Our proposed model outperforms the original MSDNet model on both the Anytime Prediction task and the Budgeted Batch Classification task.

Note that our proposed Siamese Label Auxiliary Module only assists the training of the model to improve the performance of each classifier. In the testing phase, the Siamese Label Auxiliary Module will not be used, and the output of each classifier will be directly used for prediction. The use of Siamese Label Auxiliary Module in the model does not increase the parameters of the model in the training or testing phase, but only plays an auxiliary role. The biggest feature of Siamese Label Auxiliary Module is to increase the performance of the model by means of auxiliary training without adding model parameters.

% Table generated by Excel2LaTeX from sheet 'Sheet1'
\begin{table}[htbp]
%\footnotesize
\scriptsize
  \centering
  \caption{Top-1 accuracy (\%), Top-5 accuracy (\%), BEST accuracy (\%) and negative log-likelihood (NLL) on the CIFAR-100 dataset. IND means independent training without DML or SiLa.}
\renewcommand{\arraystretch}{0.3}
    \begin{tabular}{lclllll}
    \toprule
          &       &       & Top-1 & Top-5 & BEST  & NLL \\
    \midrule
    MobileNet & IND   &       & 66.90 & 88.50 & 67.18 & 1.46 \\
    \midrule
    \multicolumn{1}{l}{\multirow{6}[12]{*}{\makecell[l]{MobileNet+\\MobileNet}}} & \multicolumn{1}{c}{\multirow{2}[4]{*}{SiLa }} & Net1  & \textbf{69.18} & \textbf{89.91} & \textbf{69.45} & \textbf{1.23} \\
\cmidrule{3-7}          &       & Net2  & \textbf{69.01} & \textbf{89.95} & \textbf{69.30} & \textbf{1.23} \\
\cmidrule{2-7}          & \multicolumn{1}{c}{\multirow{2}[4]{*}{DML}} & Net1  & 69.21 & 89.92 & 69.43 & 1.22 \\
\cmidrule{3-7}          &       & Net2  & 69.11 & 90.13 & 69.35 & 1.21 \\
\cmidrule{2-7}          & \multicolumn{1}{c}{\multirow{2}[4]{*}{SiLa+DML}} & Net1  & \textbf{70.36} & \textbf{90.92} & \textbf{70.64} & \textbf{1.13} \\
\cmidrule{3-7}          &       & Net2  & \textbf{70.00} & \textbf{90.93} & \textbf{70.28} & \textbf{1.13} \\
    \midrule
    \multicolumn{1}{p{5.065em}}{ResNet18} & IND   &       & 76.45 & 92.95 & 76.64 & 0.97 \\
    \midrule
    \multicolumn{1}{l}{\multirow{6}[12]{*}{\makecell[l]{ResNet18+\\ResNet18}}} & \multicolumn{1}{c}{\multirow{2}[4]{*}{SiLa }} & Net1  & \textbf{77.85} & \textbf{93.98} & \textbf{78.08} & \textbf{0.85} \\
\cmidrule{3-7}          &       & Net2  & \textbf{77.62} & \textbf{94.08} & \textbf{77.86} & \textbf{0.86} \\
\cmidrule{2-7}          & \multicolumn{1}{c}{\multirow{2}[4]{*}{DML}} & Net1  & 77.07 & 93.62 & 77.24 & 0.87 \\
\cmidrule{3-7}          &       & Net2  & 77.46 & 93.82 & 77.65 & 0.87 \\
\cmidrule{2-7}          & \multicolumn{1}{c}{\multirow{2}[4]{*}{SiLa+DML}} & Net1  & \textbf{78.13} & \textbf{94.51} & \textbf{78.34} & \textbf{0.80} \\
\cmidrule{3-7}          &       & Net2  & \textbf{78.03} & \textbf{94.49} & \textbf{78.28} & \textbf{0.80} \\
    \midrule
    \multicolumn{1}{l}{\multirow{6}[12]{*}{\makecell[l]{MobileNet+\\ResNet18}}} & \multicolumn{1}{c}{\multirow{2}[4]{*}{SiLa }} & Net1  & \textbf{69.17} & \textbf{89.87} & \textbf{69.40} & \textbf{1.25} \\
\cmidrule{3-7}          &       & Net2  & \textbf{77.20} & \textbf{93.80} & \textbf{77.40} & \textbf{0.89} \\
\cmidrule{2-7}          & \multicolumn{1}{c}{\multirow{2}[4]{*}{DML}} & Net1  & 69.86 & 90.37 & 70.08 & 1.20 \\
\cmidrule{3-7}          &       & Net2  & 76.62 & 93.86 & 76.86 & 0.86 \\
\cmidrule{2-7}          & \multicolumn{1}{c}{\multirow{2}[4]{*}{SiLa+DML}} & Net1  & \textbf{71.51} & \textbf{91.34} & \textbf{71.64} & \textbf{1.10} \\
\cmidrule{3-7}          &       & Net2  & \textbf{76.88} & \textbf{94.07} & \textbf{77.10} & \textbf{0.83} \\
    \midrule
    \multicolumn{1}{p{5.065em}}{VGG13} & IND   &       & 72.21 & 90.06 & 72.35 & 1.28 \\
    \midrule
    \multicolumn{1}{l}{\multirow{6}[12]{*}{\makecell[l]{VGG13+\\VGG13}}} & \multicolumn{1}{c}{\multirow{2}[4]{*}{SiLa }} & Net1  & \textbf{73.21} & \textbf{90.91} & \textbf{73.35} & \textbf{1.17} \\
\cmidrule{3-7}          &       & Net2  & \textbf{73.11} & \textbf{90.53} & \textbf{73.28} & \textbf{1.18} \\
\cmidrule{2-7}          & \multicolumn{1}{c}{\multirow{2}[4]{*}{DML}} & Net1  & 73.76 & 91.70 & 73.93 & 1.08 \\
\cmidrule{3-7}          &       & Net2  & 73.84 & 91.76 & 74.04 & 1.08 \\
\cmidrule{2-7}          & \multicolumn{1}{c}{\multirow{2}[4]{*}{SiLa+DML}} & Net1  & \textbf{74.28} & \textbf{92.03} & \textbf{74.51} & \textbf{1.05} \\
\cmidrule{3-7}          &       & Net2  & \textbf{74.22} & \textbf{92.03} & \textbf{74.41} & \textbf{1.05} \\
    \midrule
    \multicolumn{1}{l}{\multirow{6}[12]{*}{\makecell[l]{MobileNet+\\VGG13}}} & \multicolumn{1}{c}{\multirow{2}[4]{*}{SiLa }} & Net1  & \textbf{68.88} & \textbf{89.37} & \textbf{69.12} & \textbf{1.28} \\
\cmidrule{3-7}          &       & Net2  & \textbf{73.62} & \textbf{90.98} & \textbf{73.86} & \textbf{1.15} \\
\cmidrule{2-7}          & \multicolumn{1}{c}{\multirow{2}[4]{*}{DML}} & Net1  & 69.52 & 90.22 & 69.74 & 1.23 \\
\cmidrule{3-7}          &       & Net2  & 73.76 & 92.44 & 73.96 & 1.00 \\
\cmidrule{2-7}          & \multicolumn{1}{c}{\multirow{2}[4]{*}{SiLa+DML}} & Net1  & \textbf{70.74} & \textbf{91.13} & \textbf{70.93} & \textbf{1.13} \\
\cmidrule{3-7}          &       & Net2  & \textbf{74.28} & \textbf{92.78} & \textbf{74.42} & \textbf{0.96} \\
    \bottomrule
    \end{tabular}%
  \label{tab:cifar100_mutual}%
\end{table}%

\section{Experiments}
\label{experiments}
In this experiment, we will apply the SiLa Module to the training process of common networks and compare with DML. In addition, the SiLa Module is also used in the dynamic neural network and compared with MSDNet.

{\bf Datasets}.\quad Four classification datasets were used in the experiments. The {\bf MINIST} \cite{lecun1998gradient} is a dataset of handwritten digits, with a total of 10 classes and 70,000 images, of which 60,000 are used for training and 10,000 are used for testing, each with a size of $28\times  28$. Since the MINIST dataset is very simple, we only use it to visualize the training effect of the Siamese Label Auxiliary Module in our experiments. The {\bf CIFAR-10} and {\bf CIFAR-100} \cite{krizhevsky2009learning} datasets have 10 and 100 classes, respectively, and both have 60,000 images, of which 50,000 are used for training and 10,000 are used for testing, each with a size of $32\times 32$. Due to the limitation of computational resources, we did not use the ImageNet-1k dataset \cite{deng2009imagenet}, but a subset of it. The {\bf ImageNet-100} dataset is obtained by taking the first hundred classes of The ImageNet-1k dataset sorted by class name, and it contains 129,395 images for training and 5,000 images for testing. In the experiment, each image was resized to $256\times 256$ pixels, and then the center cropped to $224\times 224$ pixels.

\subsection{Results on CIFAR-100 and ImageNet-100}
\label{experiment_collaboration}
{\bf Training Details}.\quad  On the CIFAR-100 and The ImageNet-100 datasets, all models are trained for 260 epochs using stochastic gradient descent (SGD), batch size is 128, initial learning rate is 0.1, learning rate is adjusted using MultiStep, and milestones are [ 30, 55, 80, 105, 135, 160, 180, 200, 220, 240]. The network structure used in the experiment in this subsection is shown in Figure~\ref{siamese_sila_structure}. Network 1 and network 2 can use different models according to the needs of the experiment.

% Table generated by Excel2LaTeX from sheet 'Sheet1'
\begin{table}[htbp]
%\footnotesize
\scriptsize
  \centering
  \caption{Top-1 accuracy (\%), Top-5 accuracy (\%), BEST accuracy (\%) and negative log-likelihood (NLL) on the ImageNet-100 dataset. IND means independent training without DML or SiLa.}
\renewcommand{\arraystretch}{0.3}
    \begin{tabular}{lclllll}
    \toprule
          &       &       & Top-1 & Top-5 & BEST  & NLL \\
    \midrule
    ResNet18 & IND   &       & 79.09 & 94.63 & 79.72 & 0.83 \\
    \midrule
    \multicolumn{1}{l}{\multirow{6}[12]{*}{\makecell[l]{ResNet18+\\ResNet18}}} & \multicolumn{1}{c}{\multirow{2}[4]{*}{SiLa }} & Net1  & \textbf{80.39} & \textbf{95.31} & \textbf{80.96} & \textbf{0.73} \\
\cmidrule{3-7}          &       & Net2  & \textbf{80.47} & \textbf{95.31} & \textbf{80.82} & \textbf{0.72} \\
\cmidrule{2-7}          & \multicolumn{1}{c}{\multirow{2}[4]{*}{DML}} & Net1  & 80.24 & 95.16 & 81.04 & 0.72 \\
\cmidrule{3-7}          &       & Net2  & 80.30 & 95.19 & 81.00 & 0.72 \\
\cmidrule{2-7}          & \multicolumn{1}{c}{\multirow{2}[4]{*}{SiLa+DML}} & Net1  & \textbf{81.16} & \textbf{95.71} & \textbf{82.14} & \textbf{0.66} \\
\cmidrule{3-7}          &       & Net2  & \textbf{81.33} & \textbf{95.40} & \textbf{81.72} & \textbf{0.67} \\
    \midrule
    \multicolumn{1}{p{5.75em}}{ResNet34} & IND   &       & 80.63 & 95.10 & 80.90 & 0.84 \\
    \midrule
    \multicolumn{1}{l}{\multirow{6}[12]{*}{\makecell[l]{ResNet34+\\ResNet34}}} & \multicolumn{1}{c}{\multirow{2}[4]{*}{SiLa }} & Net1  & \textbf{81.33} & \textbf{95.32} & \textbf{81.56} & \textbf{0.77} \\
\cmidrule{3-7}          &       & Net2  & \textbf{81.61} & \textbf{95.83} & \textbf{82.29} & \textbf{0.70} \\
\cmidrule{2-7}          & \multicolumn{1}{c}{\multirow{2}[4]{*}{DML}} & Net1  & 81.35 & 96.04 & 81.88 & 0.70 \\
\cmidrule{3-7}          &       & Net2  & 81.53 & 95.94 & 82.06 & 0.70 \\
\cmidrule{2-7}          & \multicolumn{1}{c}{\multirow{2}[4]{*}{SiLa+DML}} & Net1  & \textbf{82.02} & \textbf{96.15} & \textbf{82.30} & \textbf{0.66} \\
\cmidrule{3-7}          &       & Net2  & \textbf{82.05} & \textbf{96.26} & \textbf{82.50} & \textbf{0.65} \\
    \midrule
    \multicolumn{1}{l}{\multirow{6}[12]{*}{\makecell[l]{ResNet18+\\ResNet34}}} & \multicolumn{1}{c}{\multirow{2}[4]{*}{SiLa }} & Net1  & \textbf{80.57} & \textbf{94.98} & \textbf{80.86} & \textbf{0.73} \\
\cmidrule{3-7}          &       & Net2  & \textbf{80.87} & \textbf{95.58} & \textbf{81.24} & \textbf{0.77} \\
\cmidrule{2-7}          & \multicolumn{1}{c}{\multirow{2}[4]{*}{DML}} & Net1  & 80.94 & 95.52 & 81.42 & 0.70 \\
\cmidrule{3-7}          &       & Net2  & 81.70 & 95.94 & 82.06 & 0.67 \\
\cmidrule{2-7}          & \multicolumn{1}{c}{\multirow{2}[4]{*}{SiLa+DML}} & Net1  & \textbf{81.08} & \textbf{95.70} & \textbf{81.74} & \textbf{0.67} \\
\cmidrule{3-7}          &       & Net2  & \textbf{82.14} & 95.90 & \textbf{82.44} & \textbf{0.64} \\
    \bottomrule
    \end{tabular}%
  \label{tab:image100_mutual}%
\end{table}%

In the experimental results in this subsection, Top-1 (\%), Top-5  (\%), and NLL are the top-1 test accuracy, top-5 test accuracy, and negative log-likelihood of the test set after model training converges, respectively. BEST (\%) is the best top-1 test accuracy that can be achieved during training. IND means independent training without using DML or SiLa. All these values are averaged over multiple experiments.

SiLa and DML are similar in that they both enable models to cooperate and learn together. We conducted comparative experiments on the CIFAR-100 and the ImageNet-100. Network1 and network2 use common models such as ResNet \cite{he2016deep}, VGG \cite{simonyan2014very}, MobileNet \cite{simonyan2014very}, etc., where DML+SiLa means using DML and SiLa at the same time, and hyperparameter $\beta=[1,1]$. The experimental results are shown in Table~\ref{tab:cifar100_mutual} and Table~\ref{tab:image100_mutual}. From the experimental results, we can conclude that both DML and SiLa can improve the classification performance of the model. When DML+SiLa is used, it has the best classification effect, and the Top-1 and Top-5 test accuracy of the model can be improved. It can be seen that SiLa is easy to use together with DML, so that the models can both assist and imitate each other, and jointly improve the classification performance of the model.

% Table generated by Excel2LaTeX from sheet 'Sheet3'
\begin{table}[htbp]
%\footnotesize
\scriptsize
  \centering
  \caption{Top-1 accuracy (\%), Top-2 accuracy (\%) and Top-5 accuracy (\%) on the CIFAR dataset when using AutoAugment. IND means independent training without DML or SiLa.}
\renewcommand{\arraystretch}{0.3}
    \begin{tabular}{lclll|ll}
    \toprule
          &       &       & \multicolumn{2}{c|}{CIFAR-10} & \multicolumn{2}{c}{CIFAR-100} \\
    \midrule
          &       &       & Top-1 & Top-2 & Top-1 & \multicolumn{1}{l}{Top-5} \\
    \midrule
    VGG19 & IND   &       & 94.55 & 98.30 & 73.82 & 90.10 \\
    \midrule
    \multicolumn{1}{l}{\multirow{2}[4]{*}{\makecell[l]{VGG19+\\VGG19}}} & \multicolumn{1}{c}{\multirow{2}[4]{*}{SiLa }} & Net1  & \textbf{94.77} & \textbf{98.34} & \textbf{74.12} & \textbf{90.46} \\
\cmidrule{3-7}          &       & Net2  & \textbf{94.76} & \textbf{98.40} & \textbf{74.66} & \textbf{90.76} \\
    \midrule
    \multicolumn{1}{p{7.5em}}{ResNet34} & IND   &       & 96.26 & 98.97 & 79.02 & 94.67 \\
    \midrule
    \multicolumn{1}{l}{\multirow{2}[4]{*}{\makecell[l]{ResNet34+\\ResNet34}}} & \multicolumn{1}{c}{\multirow{2}[4]{*}{SiLa }} & Net1  & \textbf{96.45} & \textbf{98.97} & \textbf{80.60} & \textbf{95.57} \\
\cmidrule{3-7}          &       & Net2  & \textbf{96.46} & \textbf{99.04} & \textbf{80.88} & \textbf{95.52} \\
    \midrule
    \multicolumn{1}{p{7.5em}}{MobileNet} & IND   &       & 92.19 & 97.60 & 70.34 & 90.81 \\
    \midrule
    \multicolumn{1}{l}{\multirow{2}[4]{*}{\makecell[l]{MobileNet+\\MobileNet}}} & \multicolumn{1}{c}{\multirow{2}[4]{*}{SiLa }} & Net1  & \textbf{92.24} & 97.57 & \textbf{71.28} & \textbf{91.09} \\
\cmidrule{3-7}          &       & Net2  & \textbf{93.71} & \textbf{98.15} & \textbf{73.62} & \textbf{92.21} \\
    \midrule
    WRN-16-10 & IND   &       & 96.66 & 99.08 & 80.25 & 95.07 \\
    \midrule
    \multicolumn{1}{l}{\multirow{2}[4]{*}{\makecell[l]{WRN-16-10+\\WRN-16-10}}} & \multicolumn{1}{c}{\multirow{2}[4]{*}{SiLa }} & Net1  & \textbf{96.99} & \textbf{99.24} & \textbf{82.81} & \textbf{96.57} \\
\cmidrule{3-7}          &       & Net2  & \textbf{96.91} & \textbf{99.18} & \textbf{83.05} & \textbf{96.46} \\
    \midrule
    \multicolumn{1}{p{7.5em}}{DenseNet121} & IND   &       & 96.04 & 98.92 & 80.72 & 95.45 \\
    \midrule
    \multicolumn{1}{l}{\multirow{2}[4]{*}{\makecell[l]{DenseNet121+\\DenseNet121}}} & \multicolumn{1}{c}{\multirow{2}[4]{*}{SiLa }} & Net1  & \textbf{96.24} & \textbf{98.92} & \textbf{81.23} & \textbf{95.86} \\
\cmidrule{3-7}          &       & Net2  & \textbf{96.50} & \textbf{99.12} & \textbf{81.25} & \textbf{95.77} \\
    \bottomrule
    \end{tabular}%
  \label{tab:cifar_autoaugment}%
\end{table}%

Furthermore, we conduct experiments combining SiLa with data augmentation method AutoAugment \cite{cubuk2018autoaugment}. network1 and network2 use ResNet \cite{he2016deep}, VGG \cite{simonyan2014very}, DenseNet \cite{huang2017densely}, WideResNet \cite{zagoruyko2016wide}, etc. The specific experimental results are shown in Table~\ref{tab:cifar_autoaugment}. The experimental results show that SiLa can still show good performance when combined with these data augmentation methods.

\begin{figure}[htbp]
\vskip 0.2in
\begin{center}
\centerline{\includegraphics[width=\columnwidth]{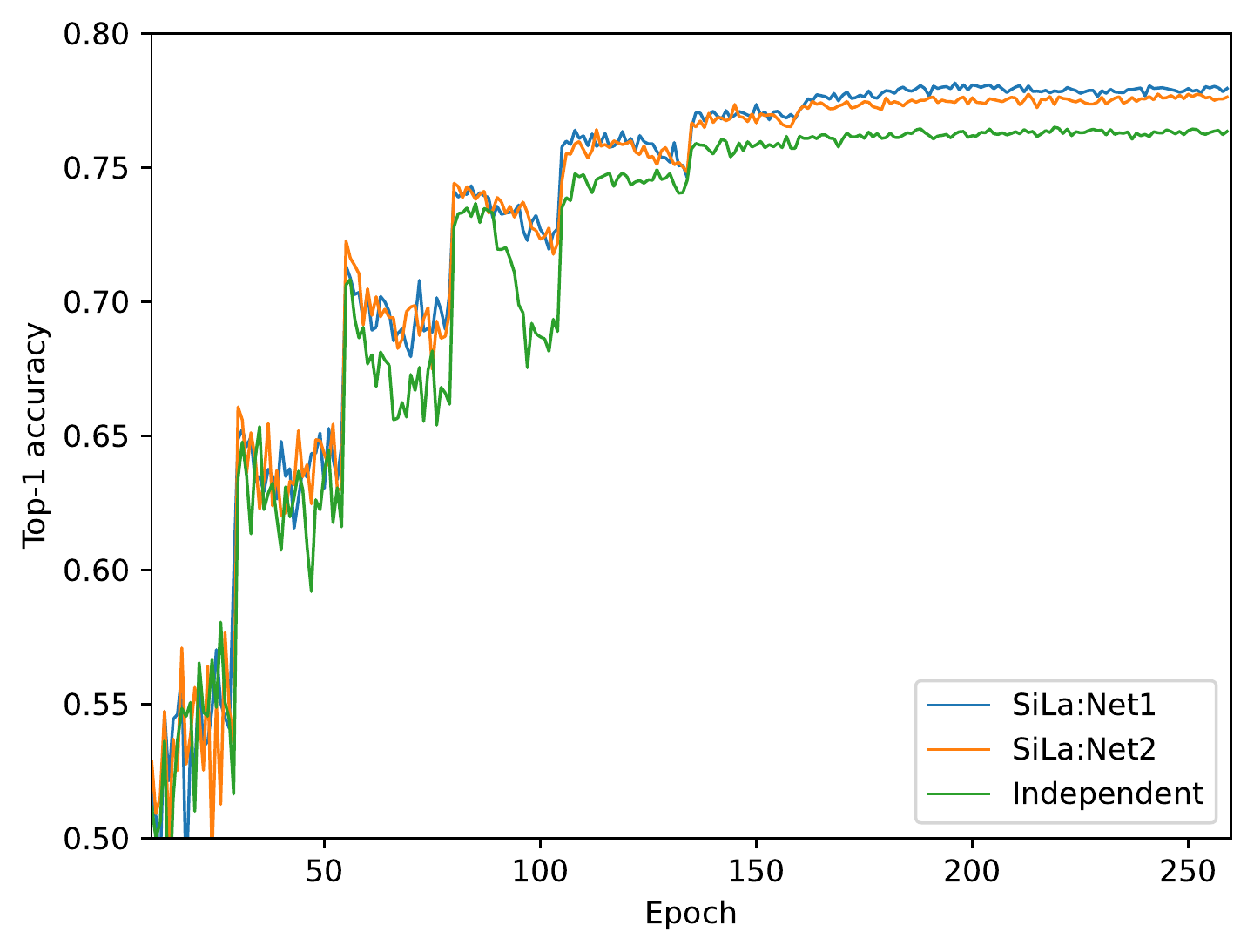}}
\caption{Top-1 test accuracy curve on CIFAR-100 dataset when ResNet18 is independently trained and using SiLa. Both Net1 and Net2 are ResNet18 when using SiLa.}
\label{resnet18_acc}
\end{center}
\vskip -0.2in
\end{figure}

We also specifically analyzed the training process of network1 and network2 using ResNet18 on CIFAR-100, and the Top-1 accuracy curve is shown in the Figure ~\ref{resnet18_acc}. It can be seen from the experimental results that using SiLa makes the model easier to train and improves the accuracy of the model. In addition, the accuracy change curves of $Net1$ and $Net2$ are very close, indicating that $Net1$ and $Net2$ assist each other and enhance each other.
In all experiments in this subsection, not only the Top-1 accuracy of the model is improved, but also the Top-5 accuracy of the model is correspondingly improved, which shows that SiLa can not only improve the accuracy of the model, but also has good robustness. NLL is the negative log-likelihood of the test set, and the reduction of NLL indicates that SiLa makes the distribution of the model output closer to the target distribution we need, which is what we need. In addition, the improvement of Bes accuracy also shows that SiLa improves the upper limit of model classification performance and can train better models.

\subsection{Dynamic Neural Networks}
\label{experiment_dnn}
{\bf Training Details}.\quad On the CIFAR-100 dataset, all models are trained for 300 epochs using stochastic gradient descent (SGD) with a batch size of 64. The initial learning rate is 0.1, divided by 10 at 150 and 255 epochs. On The ImageNet-100, all models are trained using SGD for 90 epochs with a batch size of 256. The initial learning rate is 0.1, divided by 10 at 30 and 60 epochs. In addition, the loss weight $\beta_{c}$ of each classifier is 1.

\begin{figure}[htbp]
	\centering
	\subfloat[Results on CIFAR-100]{\includegraphics[width=0.5\linewidth]{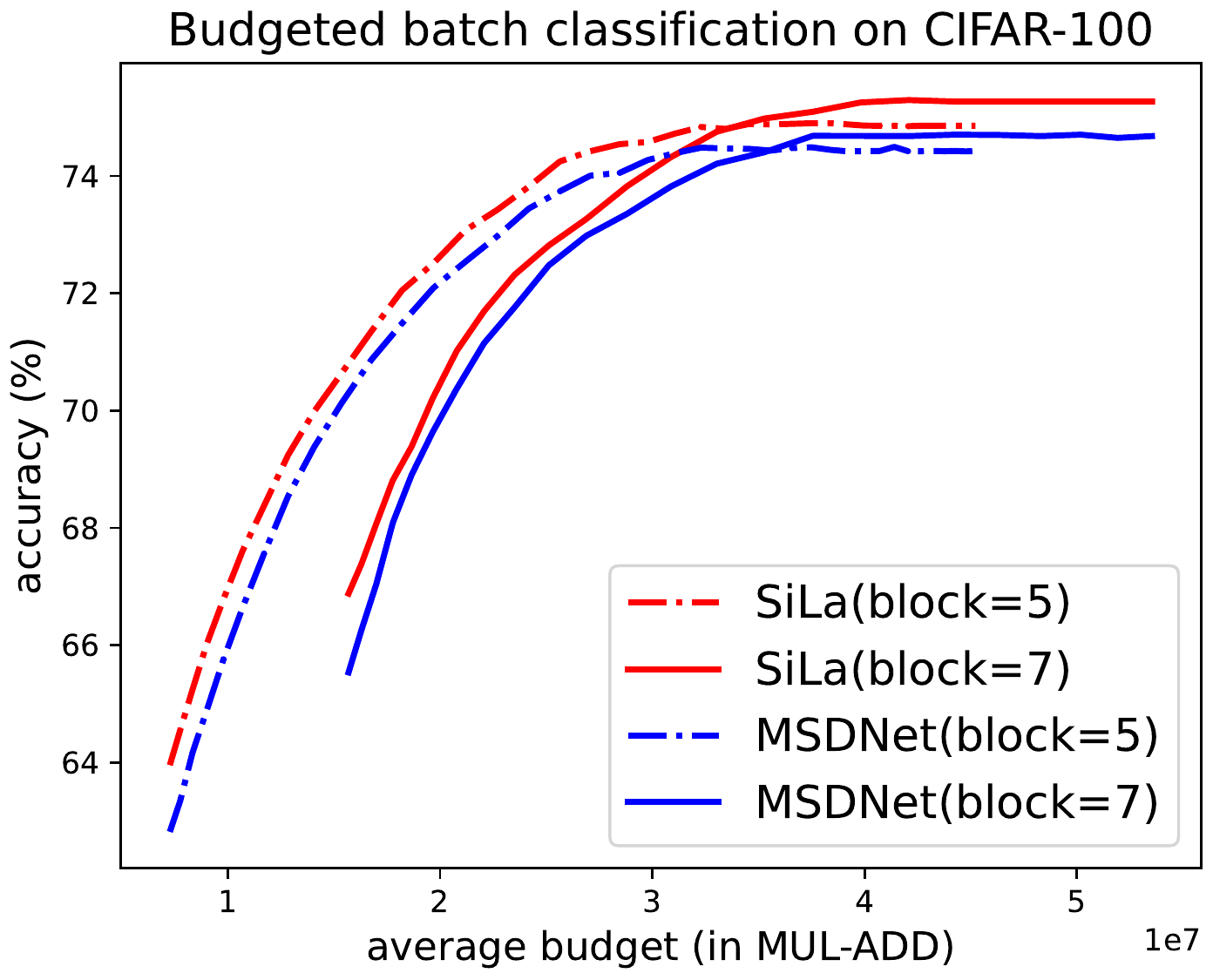}%
		\label{msd_cifar100_budget}}
	\hfil
	\subfloat[Results on ImageNet-100]{\includegraphics[width=0.5\linewidth]{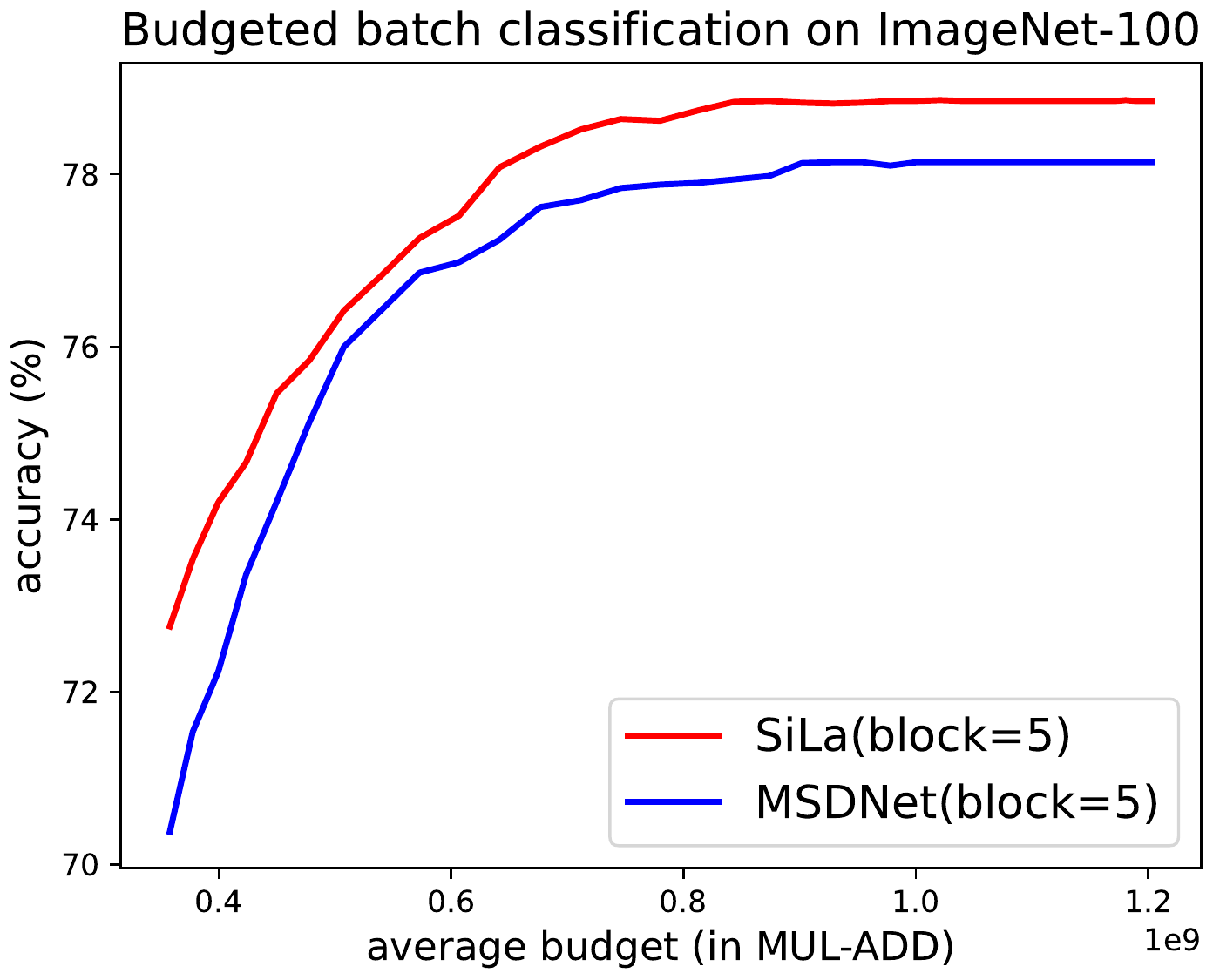}%
		\label{msd_imagenet100_budget}}
	\caption{Top-1 test accuracy results of budgeted batch classification task on CIFAR-100 (left) and ImageNet-100 (right).}
	\label{msd_budget_batch}
	\hfill

	\subfloat[Results on CIFAR-100]{\includegraphics[width=0.5\linewidth]{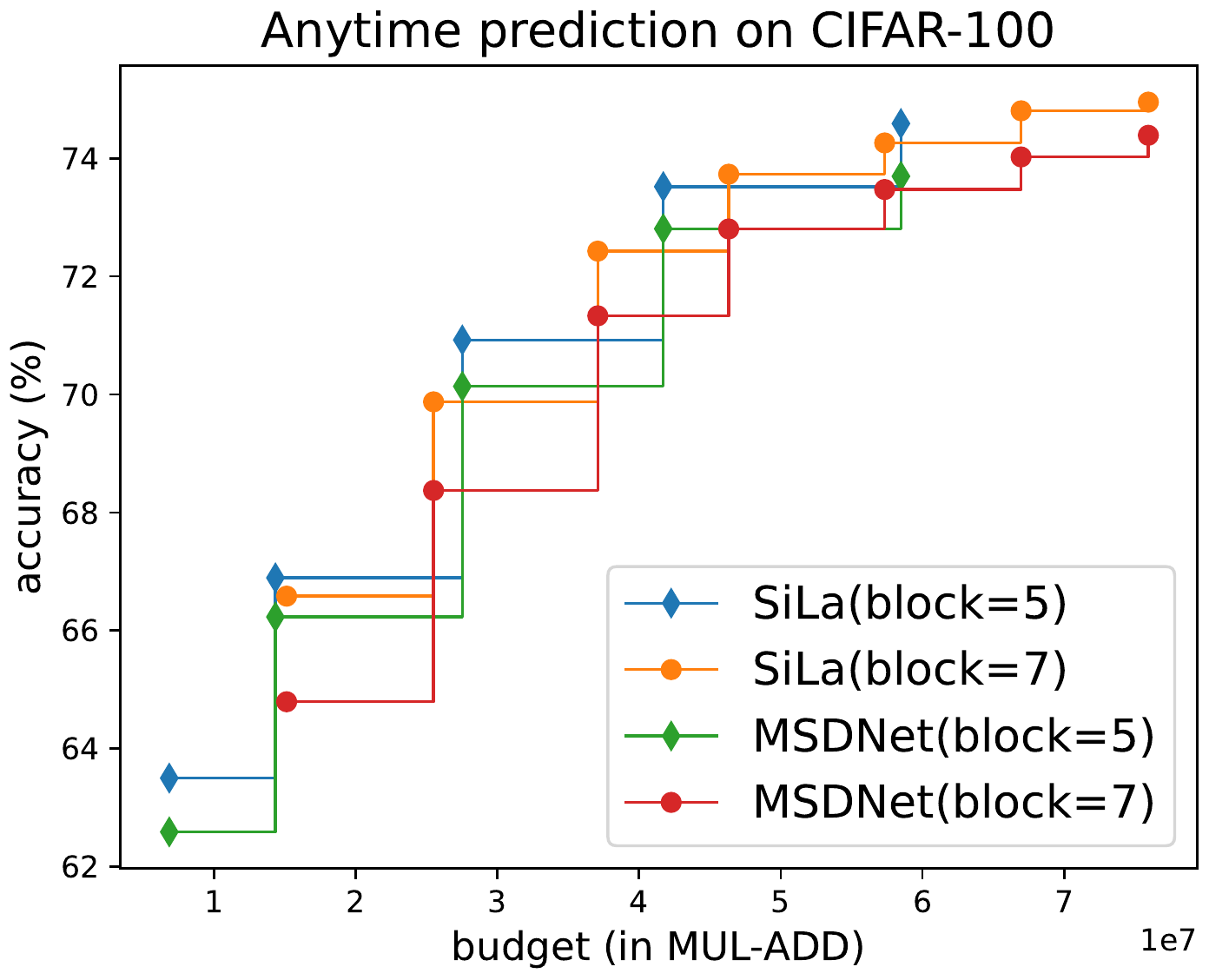}%
		\label{msdcifar100_anytime}}
	\hfil
	\subfloat[Results on ImageNet-100]{\includegraphics[width=0.5\linewidth]{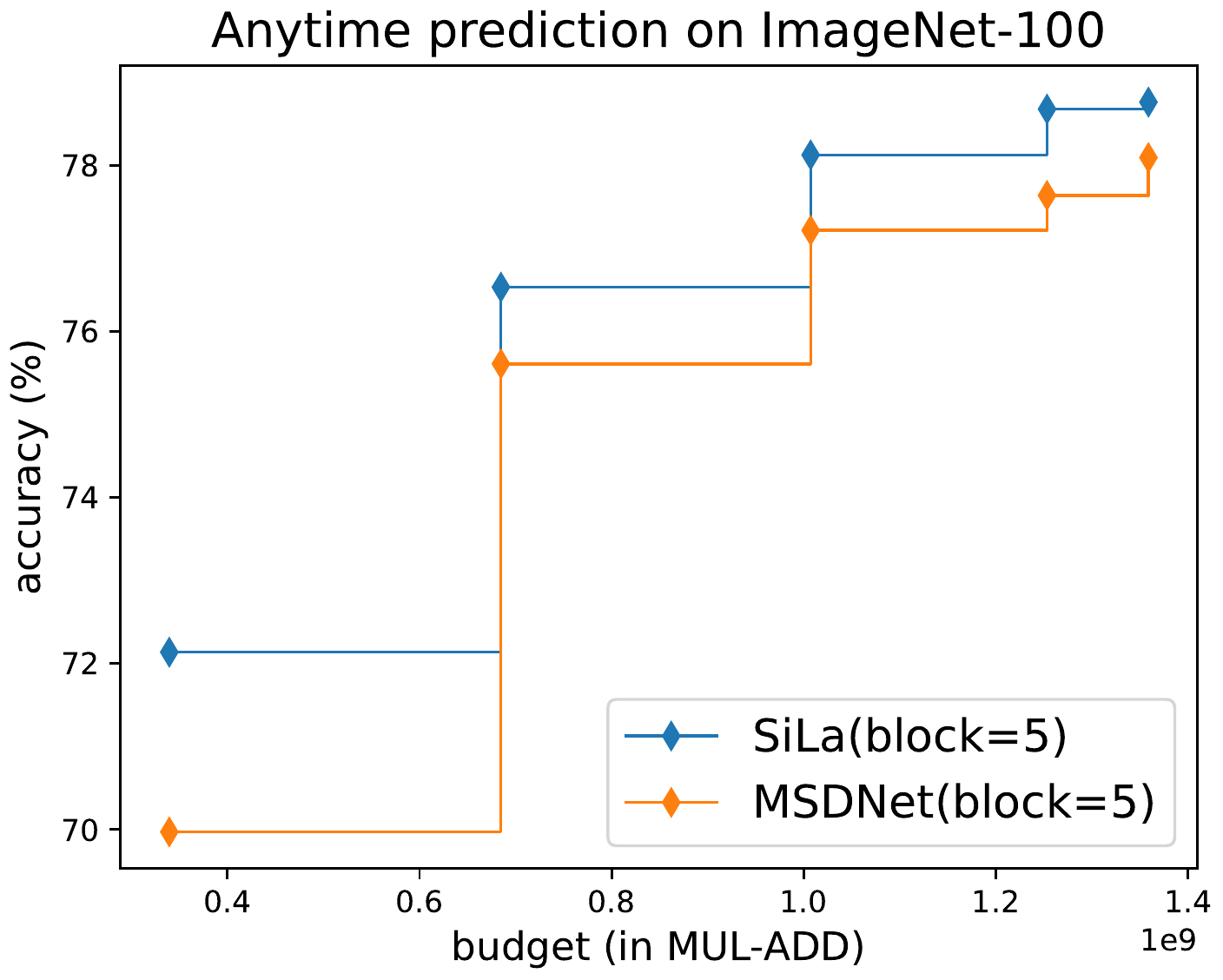}%
		\label{msdimagenet_anytime}}
	\caption{Top-1 test accuracy results of anytime prediction task on CIFAR-100 (left) and ImageNet-100 (right).}
	\label{msd_anytime}
\vskip -0.2in
\end{figure}

SiLa Module can be used not only in Collaborative Learning \cite{batra2017cooperative} with multiple independent models, but also in Dynamic Neural Networks \cite{han2021dynamic} with multiple output nodes. As shown in Figure~\ref{SilaNet_figure}, in this subsection, the network structure we use is an improvement based on MSDNet \cite{huang2018multi}. So in this experiment, we will compare with MSDNet. It should be noted that with the same number of MSD-blocks, our model has the same number of parameters as msdnet during training and testing. The experiment uses Anytime Prediction \cite{grubb2012speedboost} and Budgeted Batch Classification \cite{huang2018multi} as the test standard. The specific results are shown in Figure~\ref{msd_budget_batch}. From the experimental results we can conclude that under the same number of MSD-blocks, training with SiLa module has better performance than the original MSDNet. The results show that SiLa improves the performance of all classifiers of the model without increasing the model parameters. Through the auxiliary learning between multiple classifiers, the model can obtain better classification accuracy under the same computing resources.

It can be analyzed from these experiments that SiLa is suitable for the auxiliary training scenarios using the output of auxiliary modules. The design of auxiliary modules is not necessarily limited to an independent network, but can also have various forms and structures.

\subsection{Analyze Why does SiLa Works}
\label{analyze_sila}
\begin{figure}[htbp]
\vskip 0in
\begin{center}
\centerline{\includegraphics[width=\columnwidth]{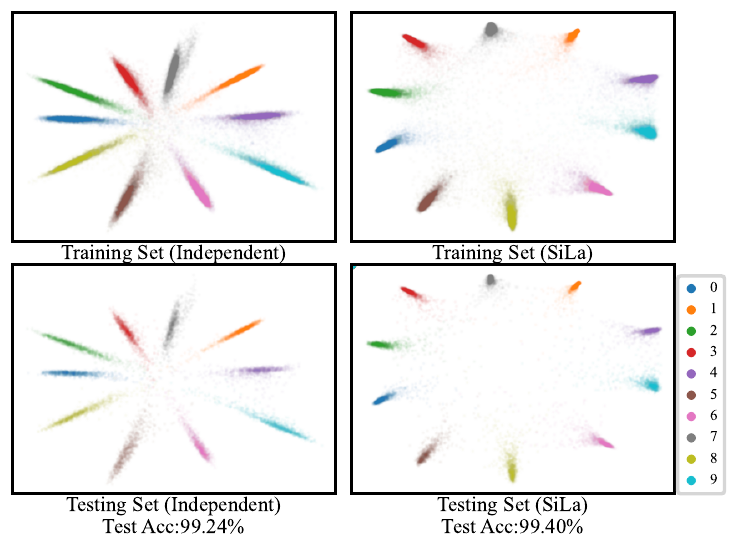}}
\caption{Feature visualization on the MNIST dataset using SiLa. Use the output of the feature layer without the fully connected layer as the visual data source}
\label{minist_sila}
\end{center}
\vskip -0.2in
\end{figure}

Referring to Large-Margin Softmax Loss \cite{liu2016large}, in order to more intuitively understand the classification effect of SiLa on the model feature layer, we conduct a simple visualization experiment on the MINIST dataset. The model was trained using SGD for 20 epochs with a fixed learning rate of 0.1 and hyperparameter $\beta=[1,1]$ . The model structure used is shown in Figure~\ref{siamese_sila_structure}. Network 1 and network 2 use the same 12-layer convolutional network. The features before the last fully-connected layer of the model are used for visualization, and the results are shown in Figure~\ref{minist_sila}, where IND means independent training without using SiLa. Since the MINIST dataset is very simple, this experiment does not focus on classification accuracy, but only for simple visualization. It can be concluded from the experimental results that the SiLa Module can make the features obtained by the model more concentrated, which makes the features easier to be classified.

\begin{figure}[htbp]
\vskip -0.2in
	\centering
	\subfloat[Training NLL Loss]{\includegraphics[width=0.48\linewidth]{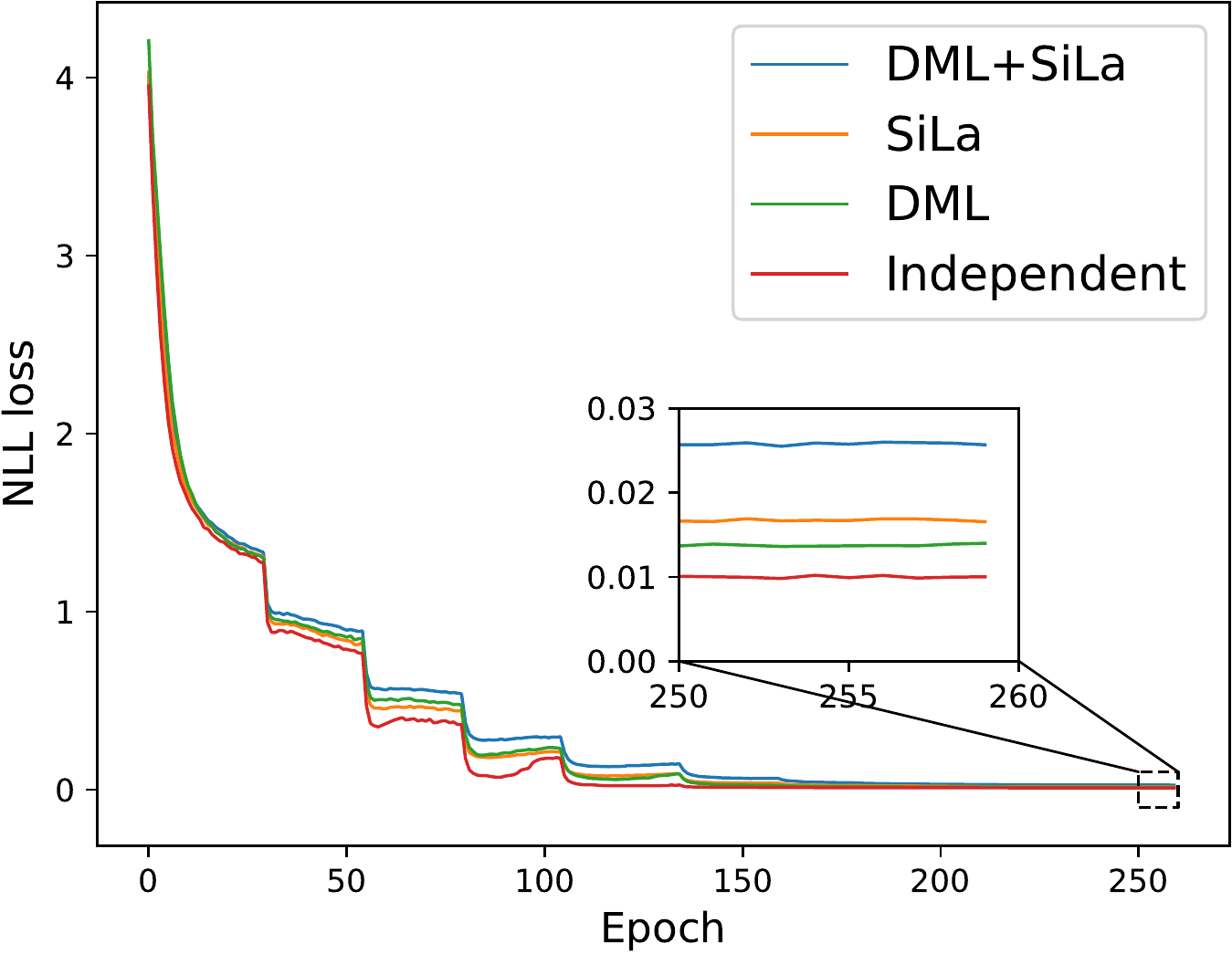}%
		\label{fig_sila_analyse_loss}}
	\hfil
	\subfloat[Training NLL Loss change]{\includegraphics[width=0.5\linewidth]{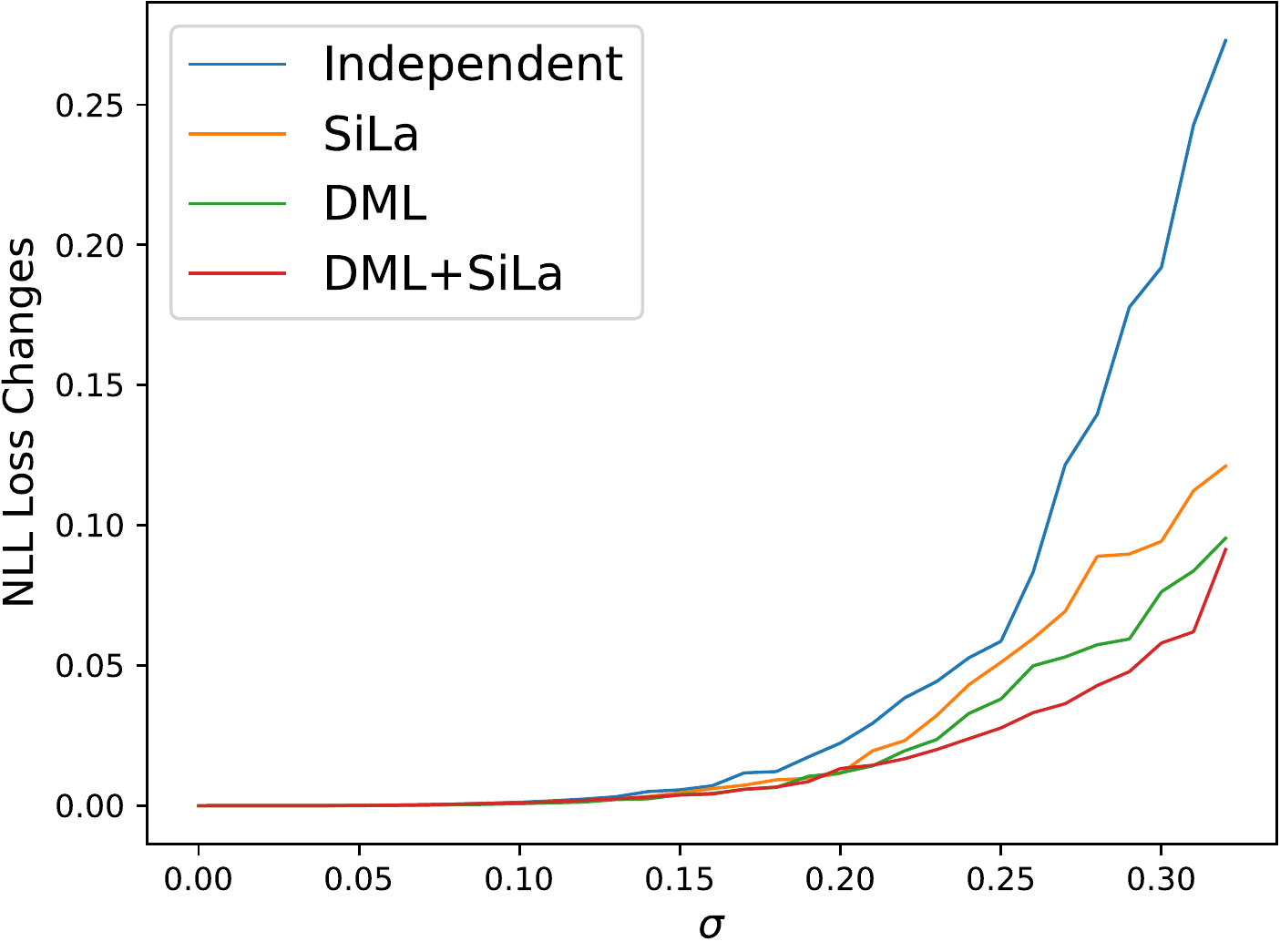}%
		\label{fig_sila_analyse_loss_change}}
	\caption{Analyze why SiLa works}
	\label{fig_sila_analyse}
\vskip -0.2in
\end{figure}

Next, we will use experiments to analyze why SiLa can improve the classification performance of the model. Analyze SiLa to help the model find parameters that make the loss smaller or more robust. We conducted experiments on the CIFAR-100 training set using the ResNet18 model. Since SiLa has different loss function structures, NLL is uniformly used for measurement in the experiment. The variation curve of training NLL Loss with epochs is shown in Figure~\ref{fig_sila_analyse_loss}. Models trained with several methods have NLL values close to 0 after convergence. But from the zoomed-in view, the model trained with SiLa has higher NLL values than other methods. This shows that the model using SiLa does not find parameters that can make the Loss smaller, but may find more robust parameters \cite{keskar2016large,pereyra2017regularizing,foret2020sharpness}. The experiments shown in Figure~\ref{fig_sila_analyse_loss_change} are used to evaluate the training NLL Loss change of the model after adding Gaussian noise with standard deviation $\sigma$ to the model parameters. We see similar results with SiLa and DML \cite{zhang2018deep}, both with smaller NLL loss changes than Independent. This shows that SiLa enables the model to find more robust parameters, thereby improving the generalization of the model \cite{chaudhari2019entropy}.

\section{Conclusion}
\label{conclusion}

In this paper, we propose an auxiliary learning method SiLa that is simple in structure and generally applicable. By using SiLa, we can improve the performance of common models without increasing the network test parameters. SiLa can also be easily combined with other Collaborative Learning methods, such as DML. Experiments show that SiLa and DML are not conflicting and have the best performance when both are used simultaneously. In addition, SiLa can also be used in early-exiting Dynamic Neural Networks with multiple output nodes, so that the model can obtain higher prediction accuracy with the same amount of computing resources. This means that SiLa can be used in various types of networks. In the future, the design and research of an auxiliary training network based on SiLa may become an interesting research point.

% In the unusual situation where you want a paper to appear in the
% references without citing it in the main text, use \nocite
%\nocite{Szegedy_2015_CVPR}

\bibliography{./paper}
\bibliographystyle{icml2021}

%%%%%%%%%%%%%%%%%%%%%%%%%%%%%%%%%%%%%%%%%%%%%%%%%%%%%%%%%%%%%%%%%%%%%%%%%%%%%%%
%%%%%%%%%%%%%%%%%%%%%%%%%%%%%%%%%%%%%%%%%%%%%%%%%%%%%%%%%%%%%%%%%%%%%%%%%%%%%%%
% DELETE THIS PART. DO NOT PLACE CONTENT AFTER THE REFERENCES!
%%%%%%%%%%%%%%%%%%%%%%%%%%%%%%%%%%%%%%%%%%%%%%%%%%%%%%%%%%%%%%%%%%%%%%%%%%%%%%%
%%%%%%%%%%%%%%%%%%%%%%%%%%%%%%%%%%%%%%%%%%%%%%%%%%%%%%%%%%%%%%%%%%%%%%%%%%%%%%%

%%%%%%%%%%%%%%%%%%%%%%%%%%%%%%%%%%%%%%%%%%%%%%%%%%%%%%%%%%%%%%%%%%%%%%%%%%%%%%%
%%%%%%%%%%%%%%%%%%%%%%%%%%%%%%%%%%%%%%%%%%%%%%%%%%%%%%%%%%%%%%%%%%%%%%%%%%%%%%%

\end{document}